\documentclass[sigconf]{acmart}

\AtBeginDocument{%
  \providecommand\BibTeX{{%
    \normalfont B\kern-0.5em{\scshape i\kern-0.25em b}\kern-0.8em\TeX}}}

\newcommand{\etal}{\emph{et al.}}
\newcommand{\eg}{\emph{e.g.}}
\newcommand{\ie}{\emph{i.e.}}
\newcommand{\etc}{\emph{etc.}}

\setcopyright{none}

\usepackage{multirow}
\usepackage{times}
\usepackage{helvet}
\usepackage{courier}
\usepackage{hyperref}       
\usepackage{url}            
\usepackage{booktabs}       
\usepackage{amsfonts}       
\usepackage{nicefrac}       
\usepackage{microtype}      
\usepackage{gensymb}
\usepackage{multirow}
\usepackage{mathtools}
\usepackage{amssymb}
\usepackage{caption}
\usepackage{subcaption}

\usepackage[symbol]{footmisc}

\begin{document}

\title{``Notic My Speech'' - Blending Speech Patterns With Multimedia}


\newcommand{\tsc}[1]{\textsuperscript{#1}} 


\author{Dhruva Sahrawat}
\authornote{Equal Contribution}
\affiliation{NUS}
\email{idsdhru@nus.edu.sg}

\author{Yaman Kumar}
\authornotemark[1]
\affiliation{IIIT-D, Adobe}
\email{yamank@iiitd.ac.in, ykumar@adobe.com}

\author{Shashwat Aggarwal}
\authornotemark[1]
\affiliation{IIIT-D}
\email{shashwata.co@nsit.net.in}

\author{Yifang Yin}
\affiliation{NUS}
\email{idsyin@nus.edu.sg}

\author{Rajiv Ratn Shah}
\affiliation{IIIT-D}
\email{rajivratn@iiitd.ac.in}

\author{Roger Zimmermann}
\affiliation{NUS}
\email{dcsrz@nus.edu.sg}



\begin{abstract}

Speech as a natural signal is composed of three parts - visemes (visual part of speech), phonemes (spoken part of speech), and language (the imposed structure). However, video as a medium for the delivery of speech and a multimedia construct has mostly ignored the cognitive aspects of speech delivery. For example, video applications like transcoding and compression have till now ignored the fact how speech is delivered and heard. To close the gap between speech understanding and multimedia video applications, in this paper, we show the initial experiments by modelling the perception on visual speech and showing its use case on video compression. On the other hand, in the visual speech recognition domain, existing studies have mostly modeled it as a classification problem, while ignoring the correlations between views, phonemes, visemes, and speech perception. This results in solutions which are further away from how human perception works. To bridge this gap, we propose a view-temporal attention mechanism to model both the view dependence and the visemic importance in speech recognition and understanding. We conduct experiments on three public visual speech recognition datasets. The experimental results show that our proposed method outperformed the existing work by 4.99\% in terms of the viseme error rate. Moreover, we show that there is a strong correlation between our model's understanding of multi-view speech and the human perception. This characteristic benefits downstream applications such as video compression and streaming where a significant number of less important frames can be compressed or eliminated while being able to maximally preserve human speech understanding with good user experience.
\end{abstract}





\settopmatter{printacmref=false}

\maketitle

\section{Introduction}
\label{intro}

Video as a medium of speech or communication delivery has grown tremendously over the past few decades. It is now being used for many aspects of day-to-day life like for online meetings, seminars, delivering lectures and entertainment applications like movies and shows. To accommodate everyone's network and bandwidth considerations, video applications have also evolved novel algorithms for compression and streaming. Yet there is much less work linking video as a computer science and multimedia construct and its linguistics perspective as a language delivery mechanism. For example, research in the linguistics domain has shown that humans do not need complete information of a word in order to recognize or understand it \cite{massaro2014speech,samuel1997lexical,samuel1996does,mcclelland1986trace}. However, the literature on perceptual video compression and streaming is yet to take rich cognitive perception information into account \cite{lee2012perceptual}. 

Speech as a medium of expression is composed of three parts- visual part (also known as visemes), aural part (also known as phonemes) and social, contextual, cultural, and other aspects (crudely the role played by language). The example of visual part of speech are lip, mouth movements and hand gestures, aural part is the phonemes, accent, intonation, \textit{etc} and language consists of concepts like structure, grammar and social context.  Consequently, three fields have developed for processing the corresponding parts of speech - visual speech recognition (or lipreading), (audio) speech recognition and natural language understanding. Even if one of the three modalities of the speech is missing, recognition efficiency and accuracy goes down. For example,  without the aural part, it is difficult to differentiate `million' from `billion', without the visual part of speech, efficiency of speech understanding goes down \cite{massaro2014speech} and without the language part, speech loses most of its communicative ability and just becomes a series of staggered words. The main idea of our work is that video as a communication medium combines all the three aspects of the speech, thus video-processing techniques should, in general, employ all the three modalities to process videos.

It is generally accepted in the psycholinguistics domain that speech is extended in time. Consequently, both articulation and understanding of a speech is affected by speech units coming before and after it \cite{mcclelland1986trace,massaro2014speech}. Secondly, over the course of a speech, a listener does not pay uniform attention to all the speech units to understand it \cite{massaro2014speech}. Thirdly, psycholinguistic phenomenon like the Ganong effect and coarticulation explain that in the cases of ambiguous speech, neighboring units help to identify the correct speech unit \cite{ganong1980phonetic,mann1980influence}.  Here, we try to model these results by a vision-speech only neural network to understand the speech perception patterns and show its application by a representative multimedia problem. We then validate the results by performing human experiments.

\textit{Visual Speech Recognition} also referred to as \textit{Lipreading}, deals with the perception and interpretation of speech from visual information alone such as the movement of lips, teeth, tongue, \etc, without using any auditory input. 
Lipreading plays a significant role in speech understanding~\cite{mcgurk1976hearing}, as it offers solutions for a multitude of applications including speech recognition in noisy environments, surveillance systems, biometric identification, and improving aids to people with hearing impairments.
Research on lipreading has spanned over centuries \cite{bulwer1648}, with modern deep learning based methods demonstrating remarkable progress towards machine lipreading \cite{assael2016lipnet,chung2016lip,chung2017lip,stafylakis2017combining}. 
Previous research on lipreading mostly ignored the closely related tasks such as modeling the correlation of different views with basic units of speech (\ie, phonemes and visemes), or identifying the phonemic or visemic subsequences that are the most important in the recognition of a phrase or a sentence. 
We argue in this paper that it is crucial to model the aforementioned tasks jointly in a lipreading system. The advantages are twofold. First, the modeling of related tasks will significantly improve the speech recognition accuracy. Second, the learnt correlations between views, phonemes, visemes, and speech perceptions will direct the attention of language learners towards the more relevant content, which provides valuable contextual information for improving the user experience in downstream video applications~\cite{christiansen2001connectionist,lake2017building,storrs2019deep}. For example, in a video conferencing or streaming system where frames have to be dropped due to poor network connections, the correlations we learnt between views/speech and human perceptions can facilitate the frame selection to achieve a better user experience than randomly dropping frames.

However, it is a highly challenging task to model the correlations between views, speech, and human perceptions. 
Visemes are the visually distinguishable units of sound, produced by different mouth actuations, such as the movements of lips, teeth, jaw and sometimes tongue \cite{fisher1968confusions}. There does not exist a one-to-one mapping between visemes and phonemes, which further makes it notoriously difficult for humans as well as machines to distinguish between different characters, words, and sentences just by looking at them. For example, different characters like \textit{p} and \textit{b} produce indistinguishable lip actuations. Similarly, phrases like \textit{wreck a nice beach} and \textit{recognize speech}, though having very different sounds and meanings, show similar visemic appearances \cite{christiansen2001connectionist}.

In this paper, we propose to model and analyse the view dependence of the basic units of speech, and the importance of visemes in recognizing a particular phrase of the English language based on attention mechanisms. We also correlate the model's understanding of view and speech with human perception and observe a strong correlation between the two components. The main contributions of our work are summarized below:

\begin{itemize}
\item To the best of our knowledge, we are the first to extend the hybrid CTC/Attention method with view-temporal attention to perform lipreading in multi-view settings. View attention learns the importance of views and temporal attention learns the importance of different video frames (viseme) for speech perception.

\item View attention dynamically adjusts the importance of each view, which helps in extracting information about the preference of model over different views for decoding of each viseme label.
\item Temporal attention reveals which video frames (viseme) are the most important in recognizing a particular phrase. It not only improves the recognition accuracy but also helps to enhance the user experience in downstream video applications from a speech perception perspective.
\item We conduct extensive experiments to compare the performance of our method with existing approaches and show an absolute improvement of 
\textbf{4.99\%} in terms of viseme error rate.
\item We discuss the problems of speech perception both from multimedia and psycholinguistics perspective. Being a new idea, we also expand on the potential applications that we see opening up at the intersection of these two areas.

\end{itemize}


\section{Related Work}
\label{rel_work}
There are several research themes in psycholinguistics which explain the human speech perception patterns.
Firstly, research suggests that speech recognition does not happen uniformly across the complete utterance. Hearing the first part of the word activates a set of words sharing the same initial part. This process continues until a specific word is identified \cite{massaro2014speech}. For example, /k\ae p/ is the beginning of both `captain' and `captive' and thus activates both the sequences until the next speech unit pins it down to one of the two choices.

Secondly, Ganong effect \cite{ganong1980phonetic} shows that in the cases of speech ambiguity, there is a tendency to perceive ambiguous speech-unit as a phoneme which would make a lexically-sound word rather than a non-sense word. An oft-quoted example is of the identification of an initial phoneme based on context when there is a ambiguity between /k/ and /g/. Humans are able to perceive /k/ as the ambiguous phoneme if the rest of the word is `iss' and /g/ if the rest is `ift'. 

Thirdly, due to the effects of co-articulation (some speech units share company much more often than others), humans adjust speech units based on the neighboring units \cite{mann1980influence}. For example, listeners perceive a /t/ followed by /s/ and /k/ followed by a /sh/ more often than vice versa \cite{elman1988cognitive}. This is the reason that when /t/ and /k/ sounds of `tapes' and `capes' were replaced with ambiguous phonemes, still the human subjects were able to associate `Christmas' (with a /s/ sound) with `tapes' (/t/ sound) and `foolish' (/sh/ sound) with `capes' (with a /k/ sound). Works such as \cite{farnetani1989cross,hoole1993comparative} indicate that this effect is cross-language and the reason for these statistical regularities is due to biological reasons \cite{lubker1982anticipatory,farnetani1999coarticulation} such as the inertia of moving speech articulators like lips and tongue.  Using this effect, some research studies have shown that by adding noise \cite{samuel1996does,samuel1997lexical} in place of some speech units, prime the humans to perceive the \textit{missing} units. All these results from the psycholinguistics and cognitive science domain have found little place in the multimedia field. Our work tries to reduce this gap so that multimedia applications could benefit from them.

The other branch of research relevant to our work is the work in visual speech in the multimedia domain. Research in automated lipreading or visual speech recognition has been extensive. Most of the early approaches consider lip reading as a single-word classification task and have relied on substantial prior knowledge bases~\cite{ngiam2011multimodal,sui2015listening}. Traditional methods such as HMMs have been utilized to perform lip reading~\cite{goldschen1997continuous}. For example, Goldschen~\etal~\cite{goldschen1997continuous} used HMMs to predict tri-viseme sequences by extracting visual features from mouth region images. Chu and Huang~\cite{chu2000} used HMMs to classify words and digit sequences. However, most of the work has focused upon predicting words or digits from limited lexicons. Further, these techniques heavily rely upon hand-engineered feature pipelines. 

With the advancements in deep learning and the increasing availability of large-scale lip reading datasets, approaches have been proposed that address lip reading using deep learning based algorithms (\eg, CNNs and LSTMs)~\cite{wand2016lipreading,petridis2018end,zhou2019modality}. Moreover, researchers have extended lip reading from single view to multi-view settings by incorporating videos of mouth section from multiple views together~\cite{lee2016multi,petridis2017end,kumar2018mylipper,kumarharnessing,uttam2019hush},. Multi-view lip reading has shown to improve performance significantly as compared to single view. Although these techniques overcome the requirement hand-engineered feature pipelines, most of the work still consider lip reading as a classification task.

Assael~\etal~proposed LipNet~\cite{assael2016lipnet}, the first end-to-end lip reading model which performs sentence-level lipreading. Several similar architectures~\cite{assael2016lipnet,chung2016lip,chung2017lip,stafylakis2017combining} have been proposed subsequently. Until recently, there have been two approaches dominant in the field of visual speech recognition, \ie, connectionist temporal classification (CTC) based and sequence-to-sequence based models respectively. To further improve the recognition accuracy, recent work have also focused upon combining CTC and attention based models together to address the task of sentence-level lip reading~\cite{xu2018lcanet,petridis2018audio}. While the literature in lip reading is vast with most of the approaches considering it as a classification task and trying to incorporate multiple views and poses together while recognizing speech, there has been very limited work on linking views with speech. 
From the perspective of human perception, it is crucial to model the view dependence in multi-view settings and identify the important phonemes and visemes in recognizing a phrase or a sentence, in order to obtain a human-like learning and thinking classifier~\cite{christiansen2001connectionist,lake2017building}.

\section{Network Architecture}
\label{sec:method}

\begin{figure}[t]
\centering
\includegraphics[width=0.47\textwidth]{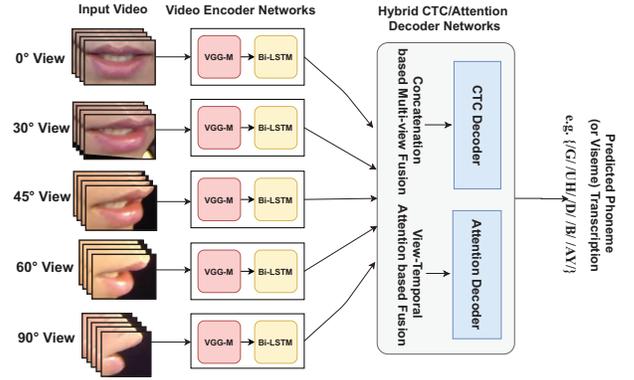}
\caption{Overview of our proposed multi-view speech recognition system.}
\label{fig:endToEndModel}
\end{figure}



Recently, a hybrid CTC/Attention architecture was proposed to solve the task of acoustic speech recognition~\cite{watanabe2017hybrid}. The hybrid mechanism combines the positives from both the approaches, \ie, forces a monotonic alignment between the input and output sequences and at the same time eliminates the assumption of conditional independence. Petridis~\etal~\cite{petridis2018audio}~employed this mechanism to perform visual speech recognition for single view settings where significant performance gain has been reported. Inspired by their work, we extend the hybrid CTC/Attention architecture to multi-view settings. Figure~\ref{fig:endToEndModel} illustrates the architecture overview of our proposed end-to-end multi-view visual speech recognition system, which consists of three major components: namely the video encoder, the view-temporal attention mechanism, and the hybrid CTC/Attention Decoder.

\subsection{Video Encoder}



The spatiotemporal video encoder maps a sequence of video frames \(x^{v} = (x_{1}^{v}, \ldots, x_{T}^{v})\) from a particular view \(v\) to a sequence of high-level feature representations \(h^{v} = (h_{1}^{v}, \ldots, h_{T}^{v})\) for that view. A separate video encoder is used for every input view \(v\). The video encoder consists of two submodules: a convolutional module and a recurrent module. 

The convolutional module is based on the VGG-M network \cite{chatfield2014return}, which is memory-efficient, fast to train, and has decent performance on ImageNet \cite{deng2009imagenet}. 
To capture the temporal dependence, we feed the output features of the convolutional module to the recurrent module of a bidirectional LSTM layer to obtain fixed-length feature representations for every input timestep. The sequence of video frames are processed by the spatiotemporal video encoder as,

\begin{equation}
    \label{eqn:video-encoder}
    \begin{split}
    &f_{t}^{v} = STCNN(x_{t}^{v})_{v}, \\
    &h_{t}^{v}, c_{t}^{v} = BiLSTM(f_{t}^{v},  h_{t-1}^{v})_{v}, \\
    \end{split}
\end{equation}
where \(x_{t}^{v}\) is the input, \(f_{t}^{v}\) is the encoded feature representation from the convolutional module and \(h_{t}^{v}\) is the fixed-length feature representation from the recurrent module at timestep \(t\) for view \(v\).

Frame-wise hidden feature representations from each view (\eg, frontal, profile, \etc) are computed by passing the video sequence for that view through the video encoder network. The encoded representations hence obtained, are then fed to the decoder networks weighted by the view-temporal attention mechanism.



\subsection{View-Temporal Attention Mechanism}
\begin{figure}[t]
\centering
\includegraphics[width=0.47\textwidth]{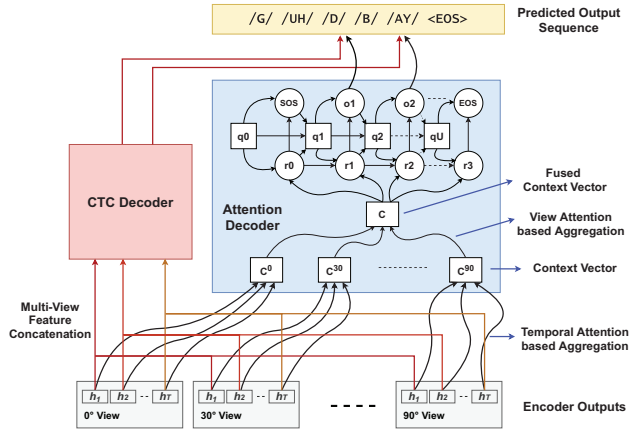}
\caption{Illustration of our proposed view-temporal attention mechanism in the lipreading system.}
\label{fig:hybridatt}
\end{figure}

We propose a view-temporal attention mechanism in the attention decoder to aggregate the features extracted from multi-view videos. As depicted in Figure~\ref{fig:hybridatt}, the temporal attention is used to compute a context vector based on the encoded representations of frames within every view, which captures an explicit alignment between the predicted output sequence and the input video frames. We experiment with three popular attention mechanism, \ie, Bahdanau or Additive Attention \cite{bahdanau2014neural}, Luong or Multiplicative Attention \cite{luong2015effective}, and Location-Aware Attention \cite{chorowski2015attention}, for the temporal attention mechanism. 

The multi-view attention is used to fuse the encoded representations from multiple views. There are two common ways to combine the encoded representations from multiple views, simple feature concatenation and weighted combination. Simple feature concatenation is a naive approach that concatenates representations from multiple views along the hidden dimension, thereby discarding information about relative significance or importance of representations from one view over another. The weighted combination based technique, on the other hand, takes into account the relative importance of different views by fusing representations from multiple views into a single representation by summing the information from individual representations based on estimated weights. We adopt the weighted combination based fusion in the attention decoder and implement it with the help of an attention mechanism that automatically learns fused representations from multiple views based on their importance. This strategy leads to enhanced representations and improved recognition accuracy based on our experimental results.

As aforementioned, we integrate the multi-view attention within the attention decoder network, where a separate temporal attention mechanism is used for every view to compute a context vector corresponding to that view. The multi-view attention is then used to fuse the context vectors from multiple views into a single fused context vector. The attention weights computed in this case change temporally, dynamically adjusting the relative importance of each view at every decoding step. Our proposed view-temporal attention mechanism is applied in the network as depicted in Figure~\ref{fig:hybridatt}, to dynamically adjust the importance of each view and extract information about the preference of the model over different views at each decoding timestep. It is also worth mentioning that we also try to integrate the multi-view attention within the encoder. However, this strategy leads to less satisfactory results as the attention weights computed for each view are fixed during the decoding of the entire output sequence.

In the CTC decoder, we combine the encoded representations from multiple views based on simple concatenation to maintain the training efficiency. The CTC decoder is not designed to include any attention mechanisms and is used as an auxiliary task to train the shared video encoder jointly with the attention decoder.

\subsection{Hybrid CTC/Attention Decoder} 
We adopt the hybrid CTC/Attention decoder to convert a stream of input sequence 
\(h = (h_{1}, \ldots, h_{T})\)
into an output sequence \(y = (y_{1}, \ldots, y_{U})\). 
The key advantages that this mechanism proffers are: (a) the CTC decoder helps in learning a monotonic alignment between the input and output feature sequences which help the model converge faster, and (b) the attention mechanism ensures that the model learns long-term dependencies among the output labels. As a shared video encoder is adopted along with two separate decoders, the CTC and attention decoders can be considered as a multi-task learning framework. The objective function optimized during the training phase is a linearly weighted combination of the CTC and attention-based losses, as stated below: 

\begin{equation}
    \label{eqn:hybrid-train}
    \begin{split}
    L_{hybrid} = &\alpha * log(p_{ctc}(y | x)) \\
    &+ (1 - \alpha) * log(p_{att}(y | x)), \\
    \end{split}
\end{equation}
where \(\alpha\) \(\in\) \((0, 1)\) is a tunable hyper-parameter. At the time of the inference, a joint CTC/Attention decoding algorithm is used to compute a joint score based on the individual probabilities from both decoders at each output timestep. The most probable hypothesis is computed as:  

\begin{equation}
    \label{eqn:hybrid-test}
    \begin{split}
    \hat{y} = \arg \max_{y \in V'} (&\lambda * log(p_{ctc}(y | x)) \\
    &+ (1 - \lambda) * log(p_{att}(y | x))), \\
    \end{split}
\end{equation}
where \(\lambda\) \(\in\) \((0, 1)\) is the weight of CTC during inference phase and \(V'\) is the augmented vocabulary, \ie, output labels plus the \([eos]\) and CTC \([blank]\) tokens, respectively.

\subsection{Implementation Details}
We implement our proposed method on the Pytorch \cite{paszke2017pytorch} framework and perform the training on a GeForce Titan X GPU with \(12GB\) of Memory. We use the ESPnet toolkit \cite{watanabe2018espnet} for training the hybrid CTC/Attention architecture. We adopt the default parameters of the ESPnet toolkit with the only exception being the CTC weight \(\alpha\) and \(\lambda\) from Equations~\ref{eqn:hybrid-train} and ~\ref{eqn:hybrid-test}. These parameters are optimized to a value of \(0.4\) and \(0.3\), respectively. All the LSTM layers have a cell size of \(256\). The output size of the network is
\(14\) for visemes, with the output vocabulary including viseme character set along with tokens for \([sos] / [eos]\) and \([blank]\).

We train the network using Adam optimizer with an initial learning rate \(lr = 0.001, \beta1 = 0.9, \beta2 = 0.999\) and with a batch size of \(16\). The network is trained for \(100\) \(epochs\) with label smoothing, dropout and early stopping applied to prevent overfitting. The weights of all the convolutional layers are initialized using the \textit{He} initializer \cite{he2015delving}. 

\begin{table*}[ht]
\small
\caption{Details of the lip reading datasets used for experiments. (\textbf{Utt.}: Utterances)}
\label{table:dataset_details}
\begin{tabular}{l|l|l|l|l|l|l}
\hline
\textbf{Dataset} & \textbf{\# Video Feeds} & \textbf{Speakers} 	& \textbf{Movement} & \textbf{Type of Videos}           																						&  \textbf{\# Utt.}	& \textbf{Vocabulary} \\ \hline
\textbf{OuluVS2} & 5                       & 53                	& None              & \vtop{\hbox{\strut Videos of simple every day phrases collected}\hbox{\strut in laboratory condition}}                 	& 	1590        			& 20                       \\ \hline
\textbf{LRW}     & 1                       &     >100           & No restriction    & \vtop{\hbox{\strut One word video segments extracted from}\hbox{\strut TED and TEDx videos }} 							&   539k        			& 	500                    \\ \hline
\textbf{LRS3}    & 1                       &   >4000            & No restriction    & \vtop{\hbox{\strut Video segments composed of long sentences}\hbox{\strut  extracted from broadcast content of BBC News }} &  	33.5k					&		 $\sim$17400       \\ \hline

\end{tabular}

\end{table*}

\section{Dataset}
\label{dataset}

We perform our experiments on three different lip reading datasets: a)~OuluVS2 \cite{anina2015ouluvs2} b)~LRW \cite{chung2016lip} c)~LRS3 \cite{afouras2018lrs3}.
The details of the dataset are give in Table \ref{table:dataset_details}. Each of the three datasets present unique challenges representative of different speech settings. 

Oulu VS2 is a clean dataset recorded in the lab settings with a few keywords. The subjects do not move their torso much and there is hardly any variation in the background, noise, \textit{etc}. This makes it suitable for studying speech patterns in a quiet room with not much environmental and other interferences. One unique advantage of Oulu VS2 is since it is recorded from multiple viewpoints, it makes it suitable for analyzing speech perception patterns from different angles.

LRW dataset is a larger dataset with approximately 500 short words. The videos contain segments of BBC News anchors articulating those words. There is a significant variation in background, speaker types, \textit{etc}. However, many a times the clips in the dataset consists of traces of the surrounding words. The videos being very short in length are good for analyzing how articulation varies in a single word. 

LRS-3 dataset includes face tracks of TEDx and TED speakers speaking longer sentences with continuous facetracks. The dataset contains many utterances with a significant vocabulary coverage. It is a good dataset to study co-articulation and other affects since the videos are longer in length. 




\section{Experiments and Results}
\label{experiments}

We perform the following set of experiments over the three public lip reading datasets:

\begin{itemize}
\item First, we empirically compare the performance of our method with some ablation models inspired by recent state-of-the-art on LRW, LRS-3 and both single and multi-view settings of OuluVS2. (Section \ref{ssec:perf_eval}).
\item Second, we evaluate the performance of our method for three different attention mechanisms, \ie, Bahdanau or Additive Attention \cite{bahdanau2014neural}, Luong or Multiplicative Attention \cite{luong2015effective}, Location-Aware Attention \cite{chorowski2015attention} (Section \ref{ssec:att_mch}).
\item Third, for every viseme, we report the view or the combination of views most significant in the detection of that viseme as found by the multi-view attention mechanism (Section \ref{ssec:diff_views}).
\item Fourth, we report the visemic subsequences which are the most important in the recognition of a phrase of the English language (Section \ref{ssec:imp_pv}).  

\item Fifth, we perform human evaluation experiments to correlate model's understanding of speech with human perception (Section \ref{ssec:human_eval}).  

\item Last, we report a set of examples on which our network performs good and bad. We further analyze certain examples on which the networks fails (Section \ref{ssec:discussion}).
\end{itemize}

\begin{figure*}[ht]%
\centering
\setlength{\belowcaptionskip}{-10pt}
\begin{subfigure}{0.33\textwidth}
	\centering
  \includegraphics[width=\textwidth]{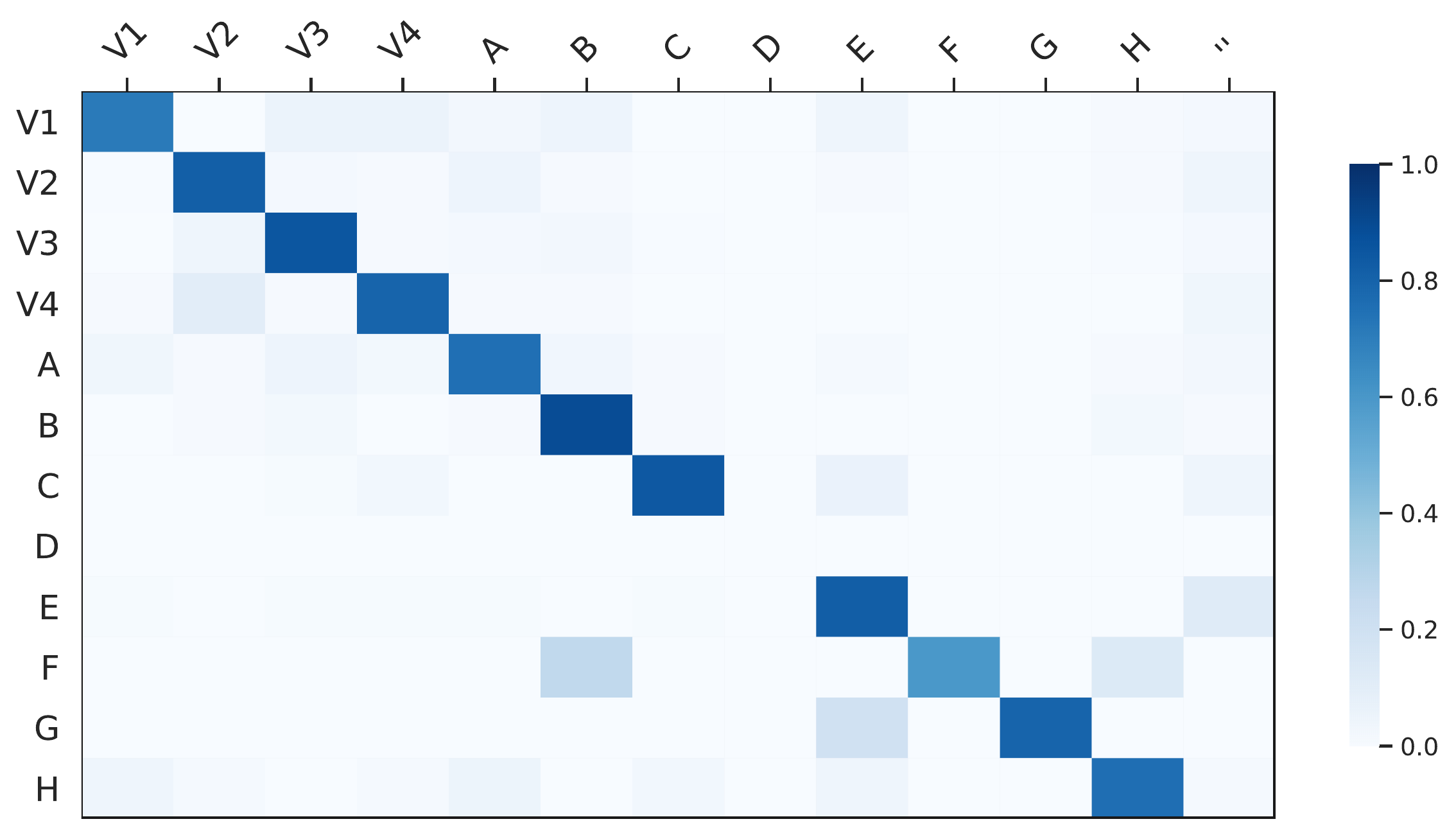}
  \caption{OuluVS2}
  \label{fig:cm1}
\end{subfigure}%
\begin{subfigure}{0.33\textwidth}
\centering
  \includegraphics[width=\textwidth]{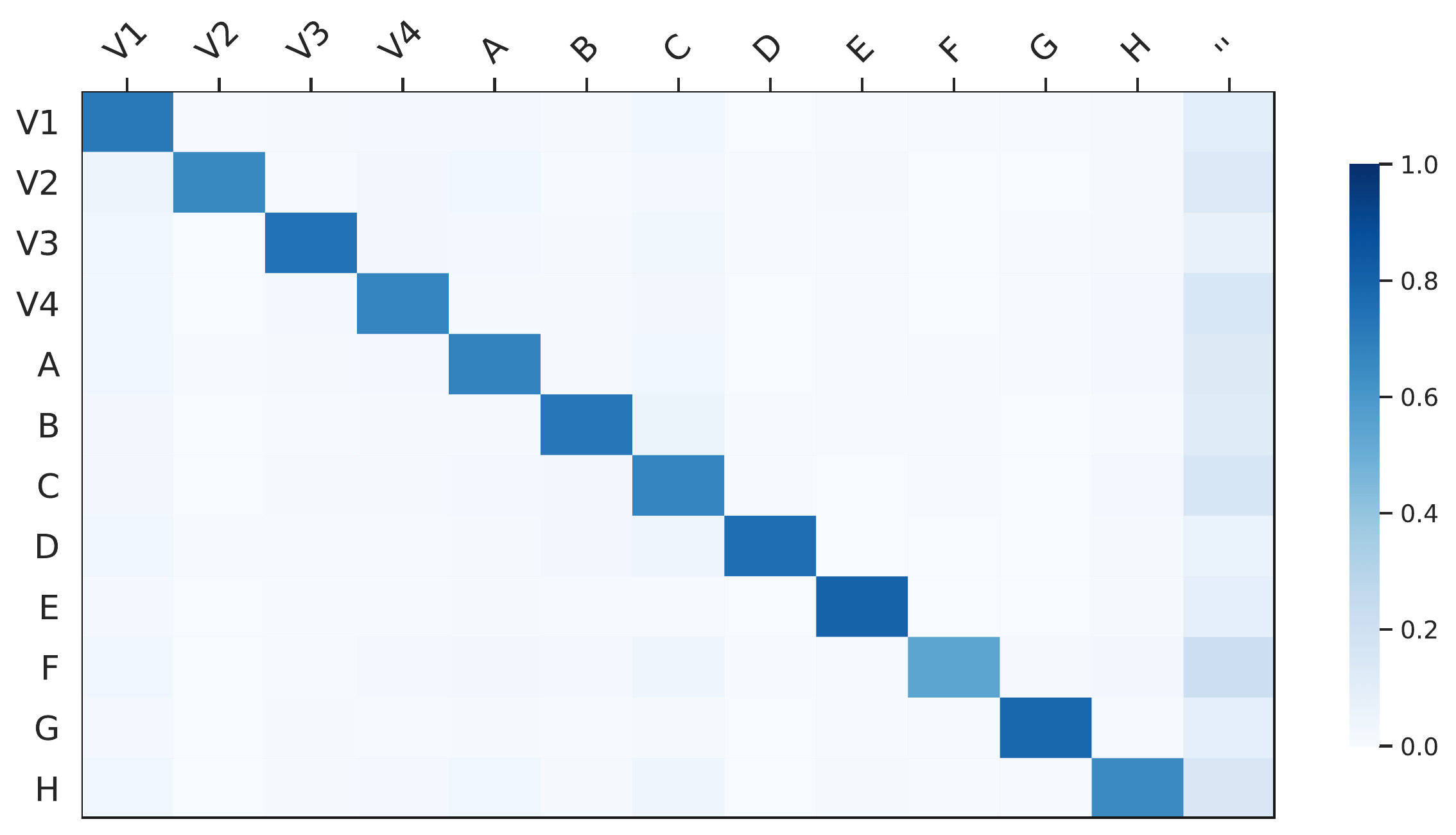}
  \caption{LRW}
  \label{fig:cm2}
\end{subfigure}
\begin{subfigure}{0.33\textwidth}
\centering
  \includegraphics[width=\textwidth]{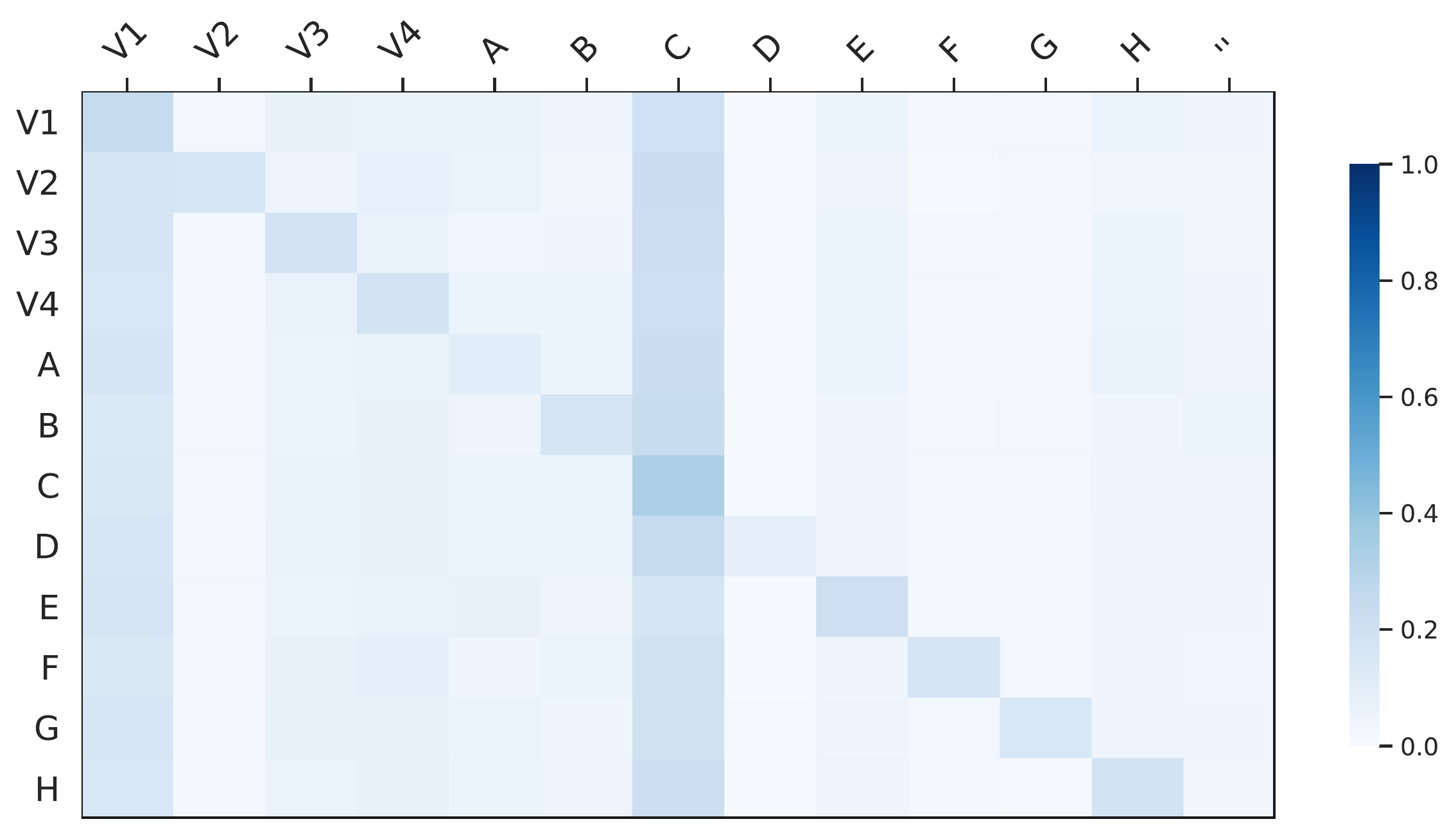}
  \caption{LRS3}
  \label{fig:cm3}
\end{subfigure}
\caption{Confusion matrix for visemes calculate separately on the test sets of following datasets: (a)OuluVS2, (b)LRW, and (c)LRS3. The row represents the actual visemes and the column represents the predicted visemes. 
}
\label{fig:confusion_matrix}
\end{figure*}

\subsection{Performance Evaluation}
\label{ssec:perf_eval}

We first perform experiments on the three datasets and show how our proposed method achieves state-of-art performance on it. 
We empirically compare the performance of our method against three ablation models inspired by recent state-of-the-art works: (a) \textbf{Baseline-CTC}: The first baseline is a CTC only method composed of STCNN with Bi-LSTMs, inspired by the method proposed in \cite{assael2016lipnet}. (b) \textbf{Baseline-Seq2Seq}: The second baseline is a sequence to sequence model without any attention mechanism, inspired by the work in \cite{chung2017lip}. (c) \textbf{Baseline-Joint-CTC/Attention}: The third baseline is a hybrid CTC/Attention architecture as employed by \cite{petridis2018audio}. The first two baselines are trained on both single-view (frontal view) and multi-view settings, while the third baseline is trained for single-view settings only. 


\begin{table}[t]
\small
\caption{Comparison with baselines on OuluVS2 dataset. Our method is in bold. (VER: Viseme Error Rate)}
\label{tab:comparison-oulu}
\begin{tabular}{l|l}
\hline
\textbf{Model} & \textbf{Viseme Error Rate} \\ \hline 
CTC only, single view & 46.41\% \\ \hline
Seq2Seq only, single view & 45.47\%  \\ \hline
CTC/Attention hybrid, single view & 18.69\%  \\ \hline
CTC only, multi view & 26.17\%\\ \hline
Seq2Seq only, multi view & 24.46\% \\ \hline
\textbf{CTC/Attention hybrid, multi view} & \textbf{13.70\%} \\ \hline
\end{tabular}
\end{table}



\begin{table}[t]
\small
\caption{Comparison with baselines on LRW and LRS3 datasets. Our method is in bold.}
\label{tab:comparison-lrw-lrs3}
\begin{tabular}{l|c|c}
\hline
\multicolumn{1}{c|}{\multirow{2}{*}{\textbf{Model}}} & \multicolumn{1}{c|}{\textbf{LRW}}               & \multicolumn{1}{c}{\textbf{LRS3}}               \\ \cline{2-3} 
\multicolumn{1}{c|}{}                                & \multicolumn{1}{c|}{\textbf{Viseme Error Rate}} & \multicolumn{1}{c}{\textbf{Viseme Error Rate}} \\ \hline

CTC only                & 48.77\%           & 92.49\%           \\ \hline
Seq2Seq only            & 33.46\%           & 76.27\%           \\ \hline
\textbf{CTC/Attention}  & \textbf{26.56\%}  & \textbf{50.84\%}  \\ \hline
\end{tabular}
\end{table}

For comparison, we compute the Viseme Error Rate (VER) as the evaluation metric. We use the beam search decoding algorithm with beam width is set to \(5\) to generate approximate maximum probability predictions from our method.
VER is defined as the minimum number of viseme insertions, substitutions and deletions required to transform the predictions from the network into the ground truth sequences, divided by the expected length of ground truth sequence. Smaller VER indicates higher prediction accuracy of the network.

Tables \ref{tab:comparison-oulu} and \ref{tab:comparison-lrw-lrs3} report the comparison results between our proposed method and the baselines on the OuluVS2, LRW and LRS-3 datasets. All results are reported for unseen speakers with both seen and unseen phrases and sentences during the training phase. The results show that our method performs better than the baselines for viseme predictions. For the OuluVS2 database, our method results in an absolute improvement of \textbf{13.46\%}and \textbf{12.47\%} in VER as compared to the CTC-only baseline and \textbf{10.76\%} in VER as compared to the Seq2Seq-only baseline, respectively. Further, our method exhibits an absolute improvement of \textbf{4.99\%} in VER as compared to the single view hybrid CTC/Attention model. The performance improvements demonstrate the importance of combining the CTC-based and Se2Seq-based architectures together. Similar gains are reported for LRW and LRS3 datasets (Table \ref{tab:comparison-lrw-lrs3}).

Single-view CTC-only and Seq2Seq models exhibits the lowest performance with very high VERs. Their multi-view counterparts, on the other hand, show significant performance improvements as compared to them, thereby confirming the importance of multi-view settings in visual speech recognition. These results are also important from cognitive science perspective and show the view dependence of visual speech.


The degradation of performance from OuluVS2 to LRW and from LRW to LRS3 can be explained by the increase in the complexity of the datasets. As shown in Table \ref{table:dataset_details}, the number of speakers, vocabulary and the dynamicity of the environment increase from OuluVS2 to LRW to LRS3. Moreover, due to resource constraints, no pretraining for LRS3 was done, which in our opinion, is the main reason for its low performance.

In Figure \ref{fig:confusion_matrix}, we present the confusion matrix for visemes calculated on OuluVS2, LRW, and LRS3 for our model. We can see that most of the visemes are confused by the visemes present in same class as them.\footnote{We include a table describing the viseme classes in the Appendix.}


\begin{table}[t]
\small
\caption{Comparison of different attention mechanisms used by our method on OuluVS2 dataset. (Att.: Attention)}
\label{tab:4}
\begin{tabular}{l|l}
\hline
\textbf{Attention Mechanism}  & \textbf{Viseme Error Rate}\\ \hline
Additive Att. & 15.12\%  \\ \hline
Multiplicative Att. & 14.20\% \\ \hline
Location-Aware Att. & \textbf{13.70\%} \\ \hline
\end{tabular}
\end{table}

\subsection{Attention Mechanisms}
\label{ssec:att_mch}


To select the best attention mechanism for the standard attention layer in our network, we evaluate the performance of our method with three popular attention mechanisms, \ie, Bahdanau or Additive Attention \cite{bahdanau2014neural}, Luong or Multiplicative Attention \cite{luong2015effective}, and Location-Aware Attention \cite{chorowski2015attention}. Additive and multiplicative attention mechanisms process the input sequence by taking into account only the content information at every timestep. Location-aware attention, on the other hand, considers both content and location information for selecting the next step in the input sequence.

In Table~\ref{tab:4}, we report VER computed by our method for each type of attention on OuluVS2\footnote{Similar results were reported on LRW and LRS3 datasets. Due to space constraints, the results are presented in the appendix.}. The results show that the location-aware attention achieves the best VER among the three attention mechanisms compared. Therefore, we adopt the location-aware attention in all the attention models in our experiments for a fair comparison.



\subsection{Importance of Different Views}
\label{ssec:diff_views}

For our third set of experiments, we analyse the weights of the multi-view attention mechanism employed in our network to find the importance of different views in the prediction of viseme labels across various output timesteps. 


\begin{table}[t]
\small
\caption{Important view for each viseme calculated on OuluVS2 dataset.}
\label{tab:5}
\centering
\begin{tabular}{l|l||l|l}
\hline
\textbf{Visemes} & \textbf{Views (in degree)} & \textbf{Visemes} & \textbf{Views (in degree)} \\ \hline
V1 & 0, 45, 60 & C & 0 \\ \hline 
V2 & 0, 60 & D & 0 \\ \hline
V3 & 0 & E & 0, 30, 45, 60, 90 \\ \hline
V4 & 0, 45 & F & 0 \\ \hline
A & 0, 45, 60 & G & 0, 30 \\ \hline
B & 0 & H & 0, 45 \\ \hline
\end{tabular}
\end{table}

\begin{figure}[htp]%
\centering
  \includegraphics[width=0.5\textwidth]{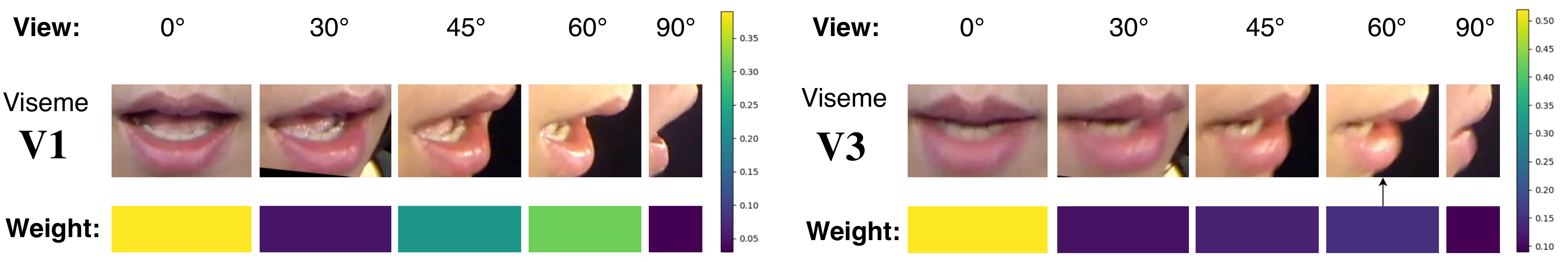}
\caption{Examples of some viseme labels along with the cropped lip section from all five views. Importance of each view is represented using the weight computed via multi-view attention mechanism.
}
\label{fig:comparisonOfDifferentViews}
\end{figure}


We report for every viseme, the average of the attention weights computed by the hierarchical attention mechanism along with a combination of views significant in the prediction of the viseme label in Table~\ref{tab:5}. We also show examples of some viseme labels along with the cropped lip section from all five views in Figure~\ref{fig:comparisonOfDifferentViews}. A reader can use these frames as reference to verify the set of views that are significant in the detection of the corresponding viseme label with those reported by our method. 
 
From Table~\ref{tab:5}, it can be observed that for most of the visemes, a single view is not enough for their detection. Most of the viseme labels require a combination of multiple views for their recognition. This is primarily due to the similar lip actuations and tongue, teeth movements between various visemes. 
Moreover, the significant views are usually a subset of the five views for the detection of a viseme, which changes dramatically from one phrase to another.



\begin{table*}[ht]
\small
\caption{Important visemes  in phrases from OuluVS2  (\textbf{1},\textbf{2}), LRW (\textbf{3}) and LRS3 (\textbf{4}). (`;':word boundary) }
\label{tab:9}
\begin{tabular}{l|l|l|l}
\hline	
\textbf{S.No.} 	& \textbf{Example Phrase} 		& 	\textbf{Visemes} 									&	\textbf{Important Visemes} 		\\ \hline
\textbf{2} 	& \textit{Goodbye} 					& 	\{H, V2, C, E, V3\} 								& 	\{V2, E, V3\} 					\\ \hline
\textbf{3} 	& \textit{Claims} 					&	\{H, A, V3, E, B\}										& 	\{H,A,V3\} 						\\ \hline
\textbf{4} 	& \textit{But it's more complicated} &	\{E, V1, C; V4, C, B; E, V1, A; H, V1, E, E, A, V1, H, V3, C, V1, C\}	& 	\{E, C; V4, B;V1,A;E,A\} \\ \hline
\end{tabular}
\end{table*}


\begin{figure*}[ht]
\centering
\includegraphics[width=0.8\textwidth]{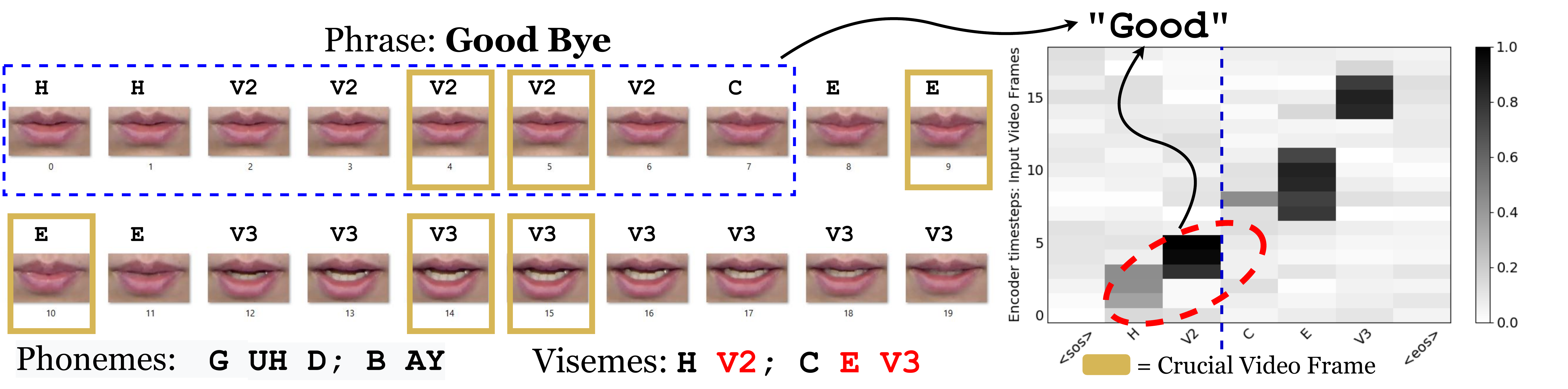}
\includegraphics[width=0.8\textwidth]{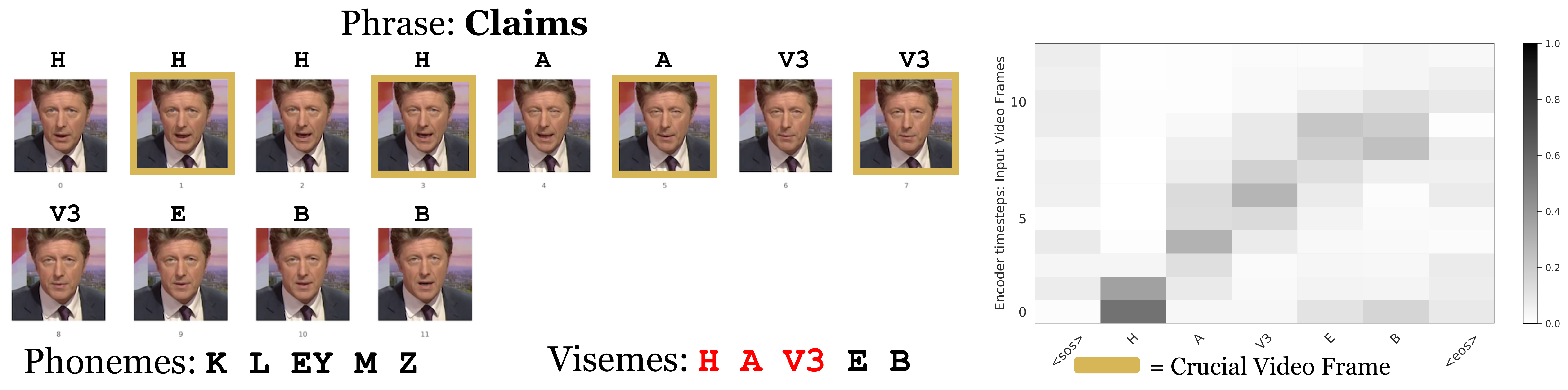}
\includegraphics[width=0.8\textwidth]{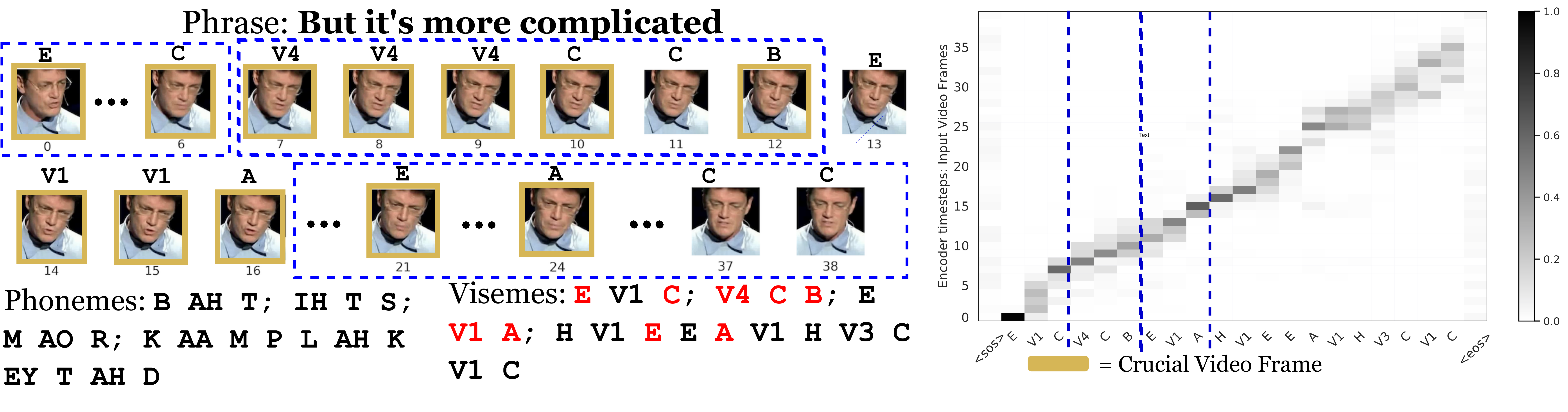}
\caption{Important visemes for phrases: \textit{“Goodbye”},\textit{“Claims”}, and \textit{“But it's more complicated”} from OuluVS2, LRW, and LRS3 respectively. The top 30\% of the frames. weighed by attention weights are considered to be important and are highlighted in yellow. Their corresponding visemes are highlighted in red. (The blue dotted lines represent word boundaries.)}
\label{fig:importantPhonemes&visemes(b)}
\end{figure*}

\subsection{Important Visemes in a Phrase}
\label{ssec:imp_pv}
We analyze the importance of visemes for recognizing a phrase by investigating the temporal attention weights obtained in our network. Based on the attention weights (by summing them up), we take the top 30\% video frames as the most important frames to understand the speech pattern.
Next, we obtain the mapping between the individual video frames and the corresponding viseme label in that frame by making a forced alignment between the audio (corresponding to the video) and the visemic transcription by using the open source tool \cite{schiel1999automatic}. Subsequently, we infer the viseme presented in a particular frame.


In Table~\ref{tab:9}, we show the most important visemes in the recognition of following four phrases:\textit{“Excuse Me”} and \textit{“Goodbye”} from OuluVS2, \textit{“Claims”} from LRW, and \textit{“But it's more complicated”} from LRS3. A detailed visualization of the video frames (frontal view) along with the viseme presented in each frame is illustrated in Figure~\ref{fig:importantPhonemes&visemes(b)} for the phrases mentioned above. The temporal attention weights are visualized on the right, and the important frames found based on the temporal attention is highlighted with yellow.

As explained in the Section~\ref{rel_work}, the first few visemes are much important for recognizing a particular word. We also observe this is true for the cases presented in the Figure~\ref{fig:importantPhonemes&visemes(b)}. Also, as was noted in \cite{massaro2014speech,mcclelland1986trace}, attention does not remain constant across the full length of the word. Rather it varies depending on how important it is for recognizing that word.


\begin{table}[ht]
\small
\caption{Human verification of videos compressed by removing unnecessary frames and by uniformally degrading the quality of video phrases.\\
\footnotesize{\normalfont Notations: The \% of people who:}\\
O-O:~{\normalfont Could not tell the difference between two original videos played side by side,} \\
M-O:~{\normalfont Thought that the that the quality of video generated by our method was at least equal to (or better than) the original,}\\ 
M-V:~{\normalfont Thought that the that the quality of video generated by our method was at least equal to (or better than) the all-frame degraded videos,}\\
V-O:~{\normalfont Thought that the quality of all-frame degraded videos were at least equal to (or better than) the original videos}
}
\label{tab:overall-human-verification-results}
\begin{tabular}{|l||l|l|}
\hline

\textbf{Video-Config} & \textbf{W/o-Audio} & \textbf{With Audio} \\ \hline

M-O & 0.4 & 0.5 \\ \hline
V-O & 0.4 & 0.3  \\ \hline
M-V & 0.8 & 0.9 \\ \hline
O-O & 0.4 & 0.6 \\ \hline

\end{tabular}
\end{table}

\begin{table}[ht]
\small
\caption{Dataset-wise human verification of videos compressed by removing unnecessary frames and by uniformally degrading the quality of video phrases.
\footnotesize \~A = Videos without Audio, A = Videos with Audio
}
\label{tab:dataset-wise-human-comparison}
\begin{tabular}{|l||l|l||l|l||l|l||}
\hline
 & \multicolumn{2}{|l||}{\textbf{LRS3}} &
 \multicolumn{2}{|l||}{\textbf{LRW}} &  
 \multicolumn{2}{|l||}{\textbf{OuluVS2}} 
 \\ \hline
\textbf{Video Config} & \textbf{A} & \textbf{\~A} & \textbf{A} & \textbf{\~A} & \textbf{A} & \textbf{\~A} \\ \hline

M-O & 0.4 & 0.3 & 0.6 & 0.6 & 0.5 & 0.3  \\ \hline
V-O & 0.3 & 0.4 & 0.3 & 0.3 & 0.2 & 0.5 \\ \hline
M-V & 0.7 & 0.4 & 0.9 & 0.9  & 0.9 & 0.8 \\ \hline
O-O & 0.6 & 0.5 & 0.8 & 0.3 & 0.2 & 0.6 \\ \hline

\end{tabular}
\end{table}

\begin{figure*}[t!]
\centering
\includegraphics[width=0.8\textwidth]{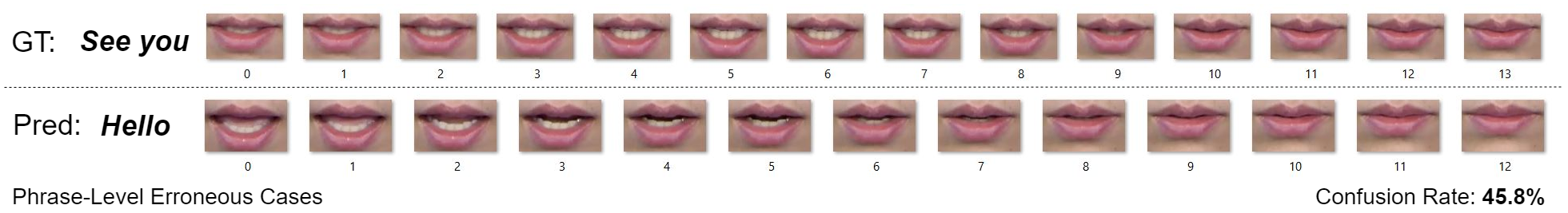}
\\
\includegraphics[width=0.8\textwidth]{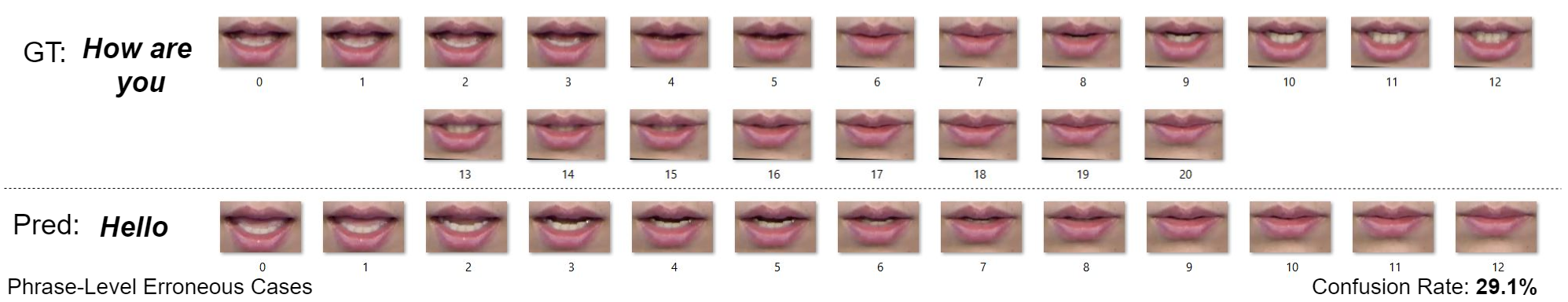}
\caption{Phrase Level erroneous examples from the dataset having a very high confusion rate. The accuracy for both of these cases is close to that for a random guess. }
\label{fig:erroneousCases2}
\end{figure*}

\subsection{Human Evaluation}
\label{ssec:human_eval}
In order to show the applicability of the results, and use them in a sample multimedia application, we perform human evaluation for the results. We take video compression as the representative multimedia application. For compressing the videos, we use an elementary compression scheme where we degrade the quality of video frames by downscaling it proportional to the inverse
of their cumulative attention weights. The resolution  $[h^{t}_{new},w^{t}_{new}]$ of a downscaled video frame $v_{new}^{t}$ is calculated using the following equation:
\[[(h^{t}_{new},w_{new})]=[\frac{h^{t}}{2}(1+ a^{t}_{cumulative}),\frac{w^{t}}{2}(1+ a^{t}_{cumulative})]\]
here h and w represent the height and width of the original video frame $v^{t}$ and $a^{t}_{cumulative}$ represents the cumulative attention weight of the of the video frame $v^{t}$. 

For a video frame, the cumulative attention weight represents the importance given by the attention model to this frame and is the sum of it's temporal attention weights corresponding to each of the viseme in the output viseme sequence.

On an average, we are able to achieve a compression factor of approximately 25\%, \ie, the size of the video get reduced by 25\%.

We compare the video compressed using the above mechanism with (1)~original videos and (2)~the videos where we uniformly degrade the quality of all the frames to achieve the same compression factor of 25\% as our method.

Next, we ask human subjects to - (1)~Given a original silent video and 4 options (2 of which are similar sounding words and the other 2 containing different sounds), identify which phrase is being spoken in that video and (2) given two videos playing side by side in sync, which one has a better quality or if both of them have the same quality

We took 3 phrases from each dataset to perform the above evaluation. The details of the phrases are given in the appendix. 

The results for these experiments are given in the Tables~\ref{tab:overall-human-verification-results} and \ref{tab:dataset-wise-human-comparison}. We also observed that human subjects were able to recognize the phrases 80\% of the time. The crucial observations from the above tables are: (1) M-V > 0.33 everywhere (randomness threshold). In fact, average M-V is 0.8 and 0.9 for non-audio and audio cases respectively. This means at least 80\% of people thought that videos generated by our method are at least as good as the videos generated by degrading the quality uniformly.
(2) If we consider O-O to be confusion metric of the annotations, M-V results are approximately 1.2 times of O-O.
(3) In most cases, M-O >= 0.4, where as V-O <= 0.4. This also means that M is preferred than V
(4) Adding audio actually increases MO as well as M-V
(5) The least O-O confusion is in OuluVS2. We believe the reason is due to high quality video and lab settings of OuluVS2.


\begin{figure}[t!]
\centering
\includegraphics[width=0.3\textwidth]{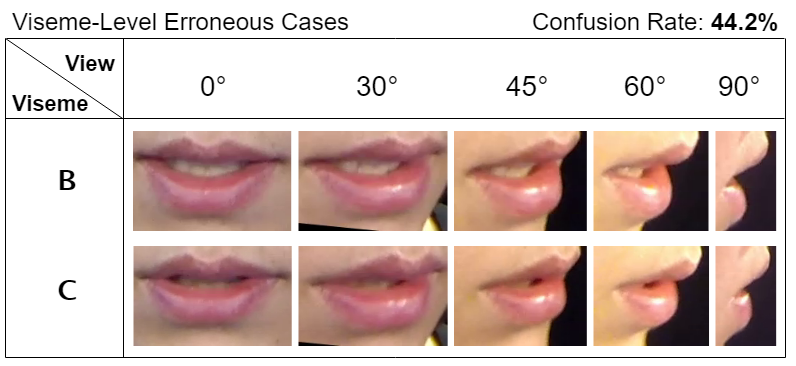}
\caption{Erroneous examples at Viseme-Level. The accuracy is close to a random guess.} 
\label{fig:erroneousCases1}
\end{figure}

\subsection{Further Discussion}
\label{ssec:discussion}

We analyse two types of error cases, (a) errors committed by our model at viseme level and (b) errors commited by our model at phrase (or sentence) level respectively. In Figure~\ref{fig:erroneousCases1}, we report a pair of visemes that our model mostly fails in discriminating them correctly. From the figure, it can be observed that it is very difficult even for a human expert to discriminate between the pair of visemes accurately. In Figure~\ref{fig:erroneousCases2}, we report some phrase-level errors committed by our model. From Figure~\ref{fig:erroneousCases2}, it can be seen that phrases such as \textit{"hello"} and \textit{"see you"} produce almost similar lip actuations and hence causing confusion in model. Also, for some phrases such as \textit{"hello"} and \textit{"how are you"}, viseme transcriptions are very similar and infact differs only at one position, which leads to added confusion in the model while decoding of such phrases. Decoding of such phrases is also largely dependant on speaker speaking characteristics, for example, a speaker with a very fast speaking rate speaking the phrase \textit{"how are you"} may sound like saying \textit{"hello"} and vice-versa.

Hence, most of the errors committed by our model are due to either indistinguishable lip actuations or underlying speaker speech characteristics which could be improved by addition audio modality along with video modality and thereby further improving the recoginition accuracy.

Our model confidently detects most of the phrases and sentences such as \textit{"excuse me"} and \textit{"good bye"}. However, there are a certain set of phrases such as \textit{"see you"} or \textit{"how are you"}, where the model fails in prediction of the phrases. The reason for this is that we take only a single part (vision) amongst the three-parts of speech. Phrases like \textit{"see you"} and \textit{"how are you"} have much in common with each other and hence are difficult to differentiate just on the basis of vision part. We will take this up as the future work. 

Another branch of research which opens up at the intersection of multimedia and psycholinguistics is how do the speech perception results vary with other important parts of speech like accent, and factors like context and environment in which a word is spoken. For example, the perception would vary with age, type of speech (example whether it is made in a movie setup or video conferencing), technical content in the speech, \textit{etc}. On the same note, this research is useful for language learners which can help them redirect their attention from the non-relevant to the relevant parts of speech \cite{osada2001strategy}. We would like to take up all these aspects as the next set of works.  

\section{Conclusion}
\label{conclusion}
We propose to improve the visual speech recognition accuracy by jointly modeling lipreading with correlated tasks in the field of cognitive speech understanding. To achieve this goal, we extend the hybrid CTC/Attention mechanism with view-temporal attention to perform lipreading in multi-view settings. Our proposed view attention module is able to learn fused representations from multiple views based on their importance and thus extract information about the preference of model over different views for decoding of each phoneme or viseme label. The temporal attention module, on the other hand, aims at finding phonemes and visemes that are the most important in the recognition of a particular phrase of the English language. We empirically compared the performance of our method with existing approaches which showed that our method obtained an absolute improvement of \textbf{4.99\%} in terms of viseme error rate. Moreover, we correlated our model's understanding of view and speech with human perception and observed a strong correlation between the two aspects. This characteristic benefits video applications such as video compression from a speech perception perspective. We also discuss the various future research directions which open up at the intersection of our work and psycholinguistics and mutlimedia. We believe that as evidenced by the multitude of applications, there is an ample potential of contributing to either of the fields.

\bibliographystyle{ACM-Reference-Format}
\bibliography{references}

\end{document}